\documentclass[10pt,twocolumn,letterpaper]{article}
\usepackage[pagenumbers]{cvpr}
\definecolor{cvprblue}{rgb}{0.21,0.49,0.74}
\usepackage[breaklinks,colorlinks,allcolors=cvprblue]{hyperref}
\hypersetup{pdftitle={Learning All-in-One Face Analysis Model from Unlabeled Videos},pdfauthor={Yuyang Sha, Pengbo Liu, Gang Luo, Ting Liu}}
\title{Learning All-in-One Face Analysis Model from Unlabeled Videos}
\newcommand{\AIOAuthorBlock}{%
\parbox{0.96\textwidth}{\centering
  {\large Yuyang Sha\textsuperscript{1,*},
  Pengbo Liu\textsuperscript{2},
  Gang Luo\textsuperscript{3},
  Ting Liu\textsuperscript{4}}\par
  \vspace{6pt}
  {\small
  \textsuperscript{1}Eye Research Center, Hangzhou Institute of Medicine, Chinese Academy of Sciences,\\
  Eye Hospital, Wenzhou Medical University, Hangzhou, 310018, China\par
  \vspace{3pt}
  \textsuperscript{2}School of Biomedical Engineering, Division of Life Sciences and Medicine,\\
  University of Science and Technology of China (USTC), Hefei, Anhui, 230026, China\par
  \vspace{3pt}
  \textsuperscript{3}Center for Artificial Intelligence Driven Drug Discovery, Faculty of Applied Sciences,\\
  Macao Polytechnic University, Macau, 999078, Macau SR\par
  \vspace{3pt}
  \textsuperscript{4}MT Lab, Meitu Inc, Beijing, 100191, China\par
  \vspace{3pt}
  \textsuperscript{*}Corresponding author: Yuyang Sha.
  }
}}
\author{\AIOAuthorBlock}

\begin{document}
\maketitle
\begin{abstract}
In this paper, we present an all-in-one facial analysis model called AIO-Face. Our model is designed to learn robust representations from unlabeled facial videos, which can be utilized for various downstream tasks such as facial landmark detection, face recognition, and facial action unit detection. The proposed method consists of three components: a feature disentanglement module, a target generation module, and a cycle-consistency regularization module. The feature disentanglement module, formed by two encoders, is able to distinguish between instance-specific features and instance-irrelevant features presented in the input facial videos. Subsequently, the target generation module produces new facial images through strategic feature combinations. To make our unsupervised learning approach more efficient, we incorporate cycle-consistency regularization. This component constrains the model to disentangle two distinct facial representations and provides additional supervision during model training. We conducted extensive experiments on various datasets for face alignment, face recognition, and facial action unit detection. The results show that our approach outperforms the evaluated unsupervised learning baselines and is comparable to some supervised approaches. Furthermore, we also evaluate the model’s performance in face reenactments, where it surpassed several leading systems.
\end{abstract}

\section{Introduction}
Face analysis is a fundamental task in computer vision, covering facial landmark detection~\cite{tourani2025landmarks}, face recognition~\cite{lafs2024, arcface}, facial action unit (AU) detection~\cite{tcae, jaanet}, and expression understanding~\cite{affectgpt2025}. These tasks describe different aspects of the same face, including identity, geometry, pose, and facial movement. Task-specific models have achieved strong results, but developing and annotating a separate model for each task remains costly. Unified models such as FaceXFormer~\cite{facexformer2025} demonstrate the value of sharing representations across facial analysis tasks. Learning transferable representations from abundant unlabeled facial videos therefore remains an important problem.

The evidence used for an assessment also depends on the application and available modalities. For example, AI-assisted screening for major depressive disorder using routine laboratory tests~\cite{laboratorymdd2026} studies laboratory and clinical indicators across multiple cohorts. This nonvisual setting illustrates the need to distinguish the information supplied by each modality from the predictions that a model has been evaluated to support.

How data are encoded also shapes the choice of learning model. MetDIT~\cite{metdit2024} transforms clinical metabolomics sequences into two-dimensional image representations and analyzes them with convolutional networks. It provides an example of adapting a representation to its input structure; the resulting images encode biochemical measurements rather than visible anatomy.

Recent advances in large language models (LLMs) and multimodal LLMs (MLLMs) have broadened the scope of machine perception. DeepSeek-R1~\cite{deepseekr12025} demonstrates the role of reinforcement-learning-based post-training in eliciting reasoning behavior from LLMs. In a specialized clinical setting, MDD-Thinker~\cite{mddthinker2026} combines supervised fine-tuning and reinforcement learning to support reasoning over clinical information for depression assessment.

In the multimodal setting, Qwen3-VL~\cite{qwen3vl2025} supports image and video understanding through a language interface, while InternVL3.5~\cite{internvl352025} investigates training and inference strategies for multimodal reasoning and efficiency. More recent omni-modal models, such as Qwen3.5-Omni~\cite{qwen35omni2026}, also incorporate audio and temporally grounded audio-visual descriptions. Together, these developments motivate studying face analysis beyond fixed-label prediction, including instruction-based interaction and natural-language descriptions of visible facial cues.

Face-centered and affective MLLMs make this connection more concrete. Face-LLaVA~\cite{facellava2026} uses face-specific instruction tuning for facial expression and attribute understanding. AffectGPT~\cite{affectgpt2025} combines multimodal evidence with descriptive supervision for emotion understanding. For clinical interviews, MIDAS~\cite{midas2026} integrates linguistic, acoustic, and visual-behavioral information for depression assessment, illustrating how visual behavior can form one component of a broader multimodal system.

These approaches also introduce different evaluation requirements: FaceXBench~\cite{facexbench2026} evaluates MLLMs across diverse face-understanding tasks and identifies substantial limitations in the models it studies. Its findings caution against treating broad visual-language capability as sufficient evidence of reliable face understanding. Precise localization, identity-related discrimination, and analysis of facial motion continue to require task-specific assessment.

Within this broader landscape, our focus is the visual representation underlying face analysis. Identity-related information should remain stable across frames, whereas pose and expression can vary over time. Unsupervised and self-supervised methods~\cite{fabnet, tcae, facecycle, pcl, fra2024} offer a way to learn such information without dense manual annotation. Explicitly separating these factors provides a concrete representation-learning objective that can be evaluated across several downstream tasks. Integration with language-conditioned models is a direction for future investigation; the present work studies this separation directly from unlabeled facial videos.

Inspired by unsupervised representation and contrastive learning, we propose a unified unsupervised face analysis model named All-In-One Face Analysis Model (AIO-Face), which can leverage unlabeled face videos for face representation learning. Our framework can disentangle the instance-specific and instance-irrelevant features from the input face videos. Specifically, the instance-specific features contain plentiful representations that are highly related to face recognition, such as shape, texture, gender, hairstyle, \textit{etc}. The instance-irrelevant features are used to describe other attributes, namely expression, and pose. The proposed AIO-Face with three components: feature disentanglement module, target generation module, and cycle-consistency regularization. The feature disentanglement module consists of two deep neural networks used to extract the instance-specific features and instance-irrelevant features from the input unlabeled face videos, respectively. The target generation module could rebuild the samples based on the extracted features from the feature disentanglement module. In order to make full use of the facial characteristics from unlabelled videos, we introduce cycle consistency regularization to enforce the consistency of the latent representation with the original one.

AIO-Face aims to build a unified model which can learn a set of robust representations from large numbers of unlabeled videos in an unsupervised manner. Furthermore, the learned representations (\textit{e.g.} instance-specific features and instance-irrelevant features) can be used for various downstream tasks in face analysis, such as facial landmark detection, face recognition, and facial AU detection. The AIO-Face can use the target generation module to generate different face images by fusing various input features. Therefore, our proposed method could also be applied to face generation or face reenactment.

In summary, the proposed AIO-Face contributes in the following three aspects:
\begin{itemize}
\item
This paper introduces a unified face analysis model, AIO-Face, based on unsupervised learning. The proposed method can learn a robust set of facial representations through deep neural networks. The AIO-Face capitalizes on the availability of unlabeled data to enhance its performance.

\item
AIO-Face is designed to learn a variety of facial representations from unlabeled videos, including instance-specific and instance-irrelevant features. These learned representations are readily applied to various facial analysis tasks, such as facial landmark detection, face recognition, facial AU detection, and face reenactment.

\item
We conduct extensive experiments to validate the proposed method in downstream tasks such as facial landmark detection, and face recognition. The results show that our method surpasses the evaluated unsupervised learning baselines and is comparable to some supervised algorithms. We also apply our model for face reenactment, and the generated results demonstrate that our model successfully disentangles the instance-specific and instance-irrelevant features.

\end{itemize}

\section{Related Work}

\textbf{Task-specific and unified face analysis.}
Facial landmark detection, recognition, and AU detection capture complementary aspects of facial appearance and movement. Representative landmark methods include SAN, Wing Loss, and FAN~\cite{san, wingloss, fan}, while AU detectors such as EAC-Net, JAA-Net, and FAT~\cite{eacnet, jaanet, fat} exploit facial regions, attention, or transformer-based feature interactions. Multi-task approaches share information across facial analysis tasks: HyperFace~\cite{hyperface} combines features from different network depths, and FaceXFormer~\cite{facexformer2025} uses a unified transformer for multiple face-perception tasks. These works establish the relevance of shared facial representations, while our study focuses on learning disentangled representations from unlabeled videos.

Portrait research also includes image-level objectives beyond facial identity and landmark localization. PortraitCraft~\cite{portraitcraft2026} provides structured composition annotations, explanations, and visual question answering for composition understanding and composition-aware generation. Its supervision complements the facial task settings above, while its evaluation objectives remain distinct from those considered in this paper.

\textbf{Unsupervised facial representation learning.}
FAb-Net~\cite{fabnet} learns facial attribute embeddings from video, and X2Face~\cite{x2face} learns a representation for driving face generation. TCAE~\cite{tcae} and FaceCycle~\cite{facecycle} use temporal or cycle-based learning signals to obtain facial representations. PCL~\cite{pcl} studies pose-disentangled contrastive learning, while FRA~\cite{fra2024} encourages consistency of global and local facial representations. LAFS~\cite{lafs2024} uses landmark-guided self-supervision for face recognition, and diffusion-based methods investigate unsupervised facial landmark and head pose discovery~\cite{tourani2025landmarks}. These approaches provide the closest representation-learning context for AIO-Face, which separates instance-specific and instance-irrelevant information and evaluates both across downstream tasks.

\begin{figure*}[t]
\centering
\includegraphics[width=\textwidth]{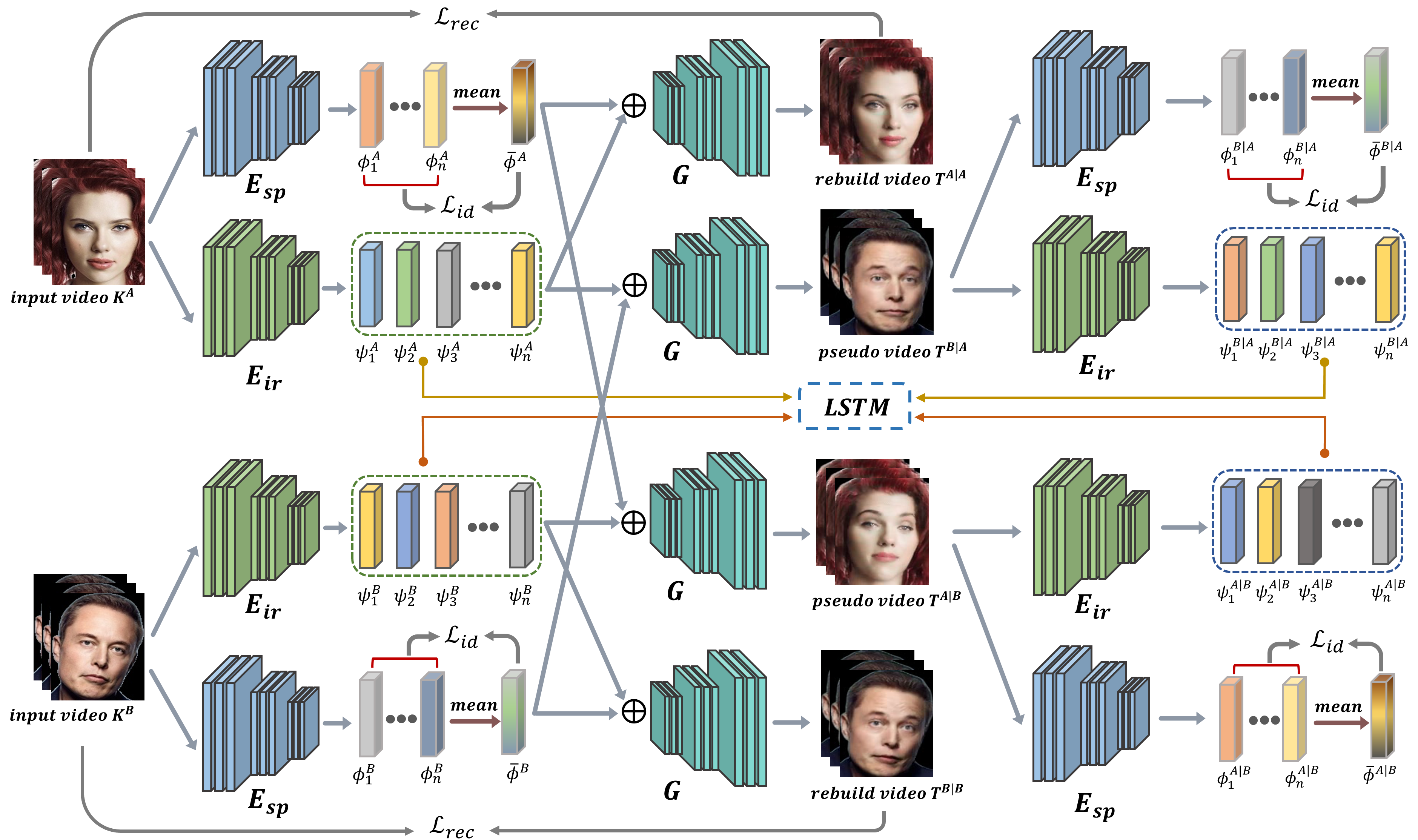}

\caption{Our AIO-Face architecture. The method contains three parts: feature disentanglement module, target generation module, and cycle-consistency regularization. The feature disentanglement module includes two encoders ($E_{sp}$ and $E_{ir}$), which are used to extract the instance-specific features and instance-irrelevant features. The target generation module (\textit{G}) attempts to generate face videos with the feature combinations of identity and attribute representations. The cycle-consistency regularization provides additional supervisions for model training.}
\label{fig:fig_2_main}
\end{figure*}

\textbf{Temporal visual feature fusion.}
Temporal integration is also studied in specialized imaging applications. CerviFusionNet~\cite{cervifusionnet2024} combines convolutional and transformer features with a GRU-based temporal module to capture changes in acetic-acid colposcopy sequences and fuse multiple image types for cervical lesion classification. This supervised medical-imaging setting illustrates temporal and multimodal feature design, with different supervision and evaluation objectives from learning facial representations from unlabeled videos.

\textbf{LLMs and domain adaptation.}
LLMs provide a language-based interface for instruction following and reasoning. DeepSeek-R1~\cite{deepseekr12025} shows how reinforcement learning can elicit reasoning behavior, motivating the study of post-training objectives alongside model scale. Domain adaptation also depends on the input representation: MDD-LLM~\cite{mddllm2025} adapts LLMs to structured clinical records represented as text for depression assessment. This setting concerns clinical information expressed through language, with different inputs and evaluation protocols from facial video representation learning.

\textbf{Multimodal foundation models.}
MLLMs extend language modeling to visual inputs. Qwen3-VL~\cite{qwen3vl2025} integrates image and video understanding with language capabilities, and InternVL3.5~\cite{internvl352025} studies staged reinforcement learning and visual-resolution and deployment strategies for reasoning and inference efficiency. Qwen3.5-Omni~\cite{qwen35omni2026} further combines text, vision, and audio and reports temporally aligned audio-visual captioning. These models support open-ended multimodal interaction, but their general-purpose evaluations do not by themselves establish performance on facial landmark localization, identity discrimination, or fine-grained facial movement analysis.

\textbf{Knowledge-guided vision-language alignment.}
For medical report generation, Sha \textit{et al.}~\cite{knowledgeguided2026} combine contrastive image-text alignment, medical knowledge injection, and retrieval of related reports with an LLM-based decoder. This provides a related example of connecting visual features to language outputs through cross-modal supervision. AIO-Face studies visual facial representations; image-text alignment and report generation are outside its evaluated tasks.

\textbf{Language-guided face and emotion understanding.}
Recent work specializes MLLMs for human-centered perception. AffectGPT~\cite{affectgpt2025} introduces descriptive emotion supervision, a multimodal model, and a benchmark designed for both categorical and free-form outputs. Face-LLaVA~\cite{facellava2026} develops face-centered instruction data and a model for facial expression and attribute understanding. These studies emphasize semantic interpretation and language interaction. AIO-Face addresses the complementary visual-learning problem of separating identity-related and time-varying facial information. Connecting such representations to MLLMs is an open direction, rather than an experimentally evaluated component of the present framework.

\textbf{Evaluation and application context.}
FaceXBench~\cite{facexbench2026} examines face understanding across multiple task categories and prompting settings, revealing limitations in the evaluated general-purpose MLLMs. Transfer to specialized applications introduces additional validation requirements. A review of medical LLMs~\cite{roadmapmed2026} discusses model foundations, clinical applications, and challenges in validation and deployment, including workflow integration. Together, these perspectives motivate matching claims to task-specific evidence. The present paper evaluates AIO-Face on visual facial analysis tasks and does not report depression screening or diagnostic experiments.

\section{Method}
Two critical challenges need to be addressed in building a unified face analysis model with unlabeled videos. The first challenge is how to effectively utilize a vast amount of unlabeled videos for model training. The second challenge involves designing a high-performance unified model that can be applied to various downstream tasks in face analysis.

Inspired by the previous works~\cite{fabnet, tcae, moco}, we introduce a novel framework joint with unsupervised and contrastive learning to address the first problem. The proposed method can take advantage of large unlabeled data for representation learning then transfer the learned features for various face analysis tasks. To solve the second challenge, the proposed method attempts to disentangle the instance-specific features and instance-irrelevant features from the input videos. These learned representations can be easily utilized for various subtasks in face analysis, such as facial landmark detection, face recognition, \textit{etc.} Therefore, our method can be applied to many downstream tasks in face analysis without designing models for each specific task. Furthermore, we also introduce the target generation module and cycle-consistency regularization into the proposed method for generating images and aligning latent features. We call the proposed method as All-In-One Face Analysis Model (AIO-Face).

\subsection{Overview}

As shown in Fig.~\ref{fig:fig_2_main}, our proposed framework contains three components: a feature disentanglement module, a target generation module, and cycle-consistency regularization. The feature disentanglement module utilizes two distinct encoders to separate the instance-specific features from those that are instance-irrelevant. The objective of this separation is to isolate the critical features for identifying a particular instance from those not directly related to the instance's identity. Next, the target generation module attempts to synthesize realistic samples with different combinations of these learned features. Furthermore, cycle-consistency regularization is employed to maintain the consistency of latent representations and ensure these facial representations can be successfully disentangled. Besides, it also can provide additional supervisory for model training.

During the training phase, we feed a pair of unlabeled videos into the proposed framework. As our model operates under an unsupervised learning framework, it does not rely on any manual labels. Consequently, we need to make full use of the structural information of the training dataset itself.
Let $\left\{ k_1^*, k_2^*, k_3^*, ..., k_n^* \right\} \in K^*$ be one of the input videos, where the $k_i^*$ represents the \textit{i-th} frame and \textit{n} denotes the total number of frames in the input video $K^*$.
The feature disentanglement module contains two encoders, referred to as instance-specific encoder (\textit{$E_{sp}$}) and instance-irrelevant encoder (\textit{$E_{ir}$}). These encoders are used to extract the instance-specific features and instance-irrelevant features from the input videos, respectively.

Once the decoupled representations are obtained, the target generation module (\textit{G}) combines the features encoded by \textit{$E_{sp}$} and \textit{$E_{ir}$}, then leverages them to generate face videos. In our approach, the target generation module produces four sequences: two of them consist of reconstructed original videos, and the remaining two sequences comprise cross-generated pseudo videos. The instance-specific features and instance-irrelevant features from different input videos are used for generating pseudo videos.

Since we lack pixel-wise labels or supervised information in the training stage, we employ the cross-verified disentangling strategy to guide the model’s learning process. The cycle-consistency regularization is introduced to ensure the quality of the generated sequences by enforcing the consistency between the input videos and their reconstructed counterparts.

\subsection{Feature Disentanglement Module}

The feature disentanglement module in AIO-Face contains two encoders, referred to as the instance-specific encoder ($E_{sp}$) and instance-irrelevant encoder ($E_{ir}$).
The $E_{sp}$ is responsible for learning the identity representations from the input face sequences, while the  $E_{ir}$ extracts features that are unrelated to identity.
During the training stage, our AIO-Face model requires two face videos, denoted as $K^A$ and $K^B$, to be fed as input. AIO-Face encodes the input videos with \textit{$E_{sp}$} and \textit{$E_{ir}$}, then gets their representations [$\Bar{\phi}^A$, $\psi^{A}_{i}$] and [$\Bar{\phi}^B$, $\psi^{B}_{i}$], respectively.

For a video containing \textbf{N} frames, the {$E_{sp}$} expects to encode a embedding $\phi_{i}^*=E_{sp}(k_i^*)$. Then we obtain the instance-specific features by taking mean average from all the learned representations, as following:

\begin{equation}
\Bar{\phi}^*=\frac{1}{N}\sum^{N}_{i=1}(E_{sp}(k_i^*)),
\end{equation}
where the $\Bar{\phi}^*$ denotes the average instance-specific features of input video $K^*$.

Each frame in the input videos contains various facial attributes, including poses and expressions. To capture these attributes independently for each frame, our model generate an embedding vector $\psi_{i}^*=E_{ir}(k_i^*)$.

In contrast, the $E_{sp}$ aims to extract instance-specific features that are similar across frames.
Therefore, we impose a constraint on the instance-specific representations to ensure their similarity to the global average representation. We define this constrain as $\mathcal{L}_{id}$ which servers constrain the instance-specific representations:

\begin{equation}
\mathcal{L}_{id} = \frac{1}{N} \sum^{N}_{i=1} \Vert \Bar{\phi}^* - \phi_{i}^* \Vert ^{2}_{2} ,
\end{equation}
where the $\Bar{\phi}^*$ and $\phi_i^*$ are the average instance-specific features and the individual ones.

Furthermore, we introduce the orthogonal loss to ensure that the instance-specific features and instance-irrelevant features are orthogonal to each other. The $\mathcal{L}_{orth}$ is defined as follows.
\begin{equation}
\mathcal{L}_{orth} = \frac{1}{N} \sum^{N}_{i=1} \Vert \phi^{*}_{i} \cdot \psi^{*}_{i} \Vert ^{2}_{2},
\end{equation}

\subsection{Target Generation Module}
To ensure that the facial representations extracted by proposed encoders ($E_{sp}$ and $E_{ir}$) contain both instance-specific and instance-irrelevant information, we feed them into the target generation module (\textit{G}) for video generation. The \textit{G} outputs four videos, including $T^{A|A}$, $T^{A|B}$, $T^{B|A}$, and $T^{B|B}$. Let's take the $T^{A|B}$ as an example, the generated video can be defined as follows:

\begin{equation}
t_{\Bar{\phi}^A, \psi^B_i} = G (\Bar{\phi}^A, \psi^B_i) \in T^{A|B} .
\end{equation}

The $t_{\Bar{\phi}^A, \psi^B_i}$ denotes the generated frame, which forms the generate video $T^{A|B}$ .The $\Bar{\phi}^A$ represents the average instance-specific features from the input video $K^A$, and the $\phi^B_i$ denotes the instance-irrelevant features from the \textit{i-th} frame input video $K^B$.
Therefore, we can control the identity and attributes in the generated face videos by selecting instance-specific features and instance-irrelevant features.
In our setting, the AIO-Face would reconstruct two original videos ($T^{A|A}$ and $T^{B|B}$) and generate two pseudo videos ($T^{A|B}$ and $T^{B|A}$). The original videos are reconstructed by the $\phi$ and $\psi$ from the same input videos. However, the pseudo videos use features from different videos.

As shown in Fig.~\ref{fig:fig_2_main}, we first employ the feature disentanglement module to get the instance-specific features and instance-irrelevant features. Then, the target generate module selects instance-specific features and instance-irrelevant features from the same videos to reconstruct two original videos and chooses these features from different videos to generate two pseudo ones. The reconstructed original videos should be the same as input ones, and we formulate the reconstruction loss as follows:
\begin{equation}
\mathcal{L}_{rec} = \Vert {K^{A}} - T^{A|A} \Vert _1 + \Vert {K^{B}} - T^{B|B} \Vert _1 ,
\end{equation}
where the $K^*$ denotes the input video, and $ T^{*|*}$ is the reconstructed original video.

\subsection{Cycle-Consistency Regularization}
The performance of AIO-Face in feature disentanglement can be inferred from the quality of the generated videos. We expect that the disentangled facial representations should be cycle consistent. Therefore, we design introduce the cycle-consistency regularization, which provides additional supervision to facilitate the disentanglement of identity and attribute representations.
After the target generator module, we could get four generated face videos, where two of them are reconstructed videos and the rest are cross-generated pseudo samples. All pseudo videos are generated by instance-specific features and instance-irrelevant features from different input videos.

Let us take the generated pseudo video $T^{A|B}$ to illustrate the concept. If the instance-specific encoder only encodes the instance-specific information, the instance-specific features $\Bar{\phi}^A$ are expected to be the same as $\Bar{\phi}^{A|B}$. Since the pseudo video $T^{A|B}$ is generated by concatenating the features of  [$\Bar{\phi}^A$, $\psi^B$], and the identity representation contained in the generated pseudo video should be identical to the original one.

We formulate a cycle consistent loss to minimize the distance between the features extracted from the original and pseudo video, which is defined as follows:

\begin{equation}
\mathcal{L}_{cyc}^{sp} = \Vert \Bar{\phi}^A - \Bar{\phi}^{A|B} \Vert _2^2 + \Vert \Bar{\phi}^B - \Bar{\phi}^{B|A} \Vert _2^2 .
\end{equation}

The instance-irrelevant features involved in $T^{A|B}$ are expected to be similar to $K^B$. The instance-irrelevant features $\psi^{A|B}$ in pseudo video should be consistent with $\psi^{B}$ in the input samples. We minimize the discrepancy of the embedding for facial attributes by:

\begin{equation}
\mathcal{L}_{cyc}^{ir} = \sum _{i=1}^N \Vert \psi^A_i - \psi^{B|A}_i \Vert _2^2 + \Vert \psi^B_i - \psi^{A|B}_i \Vert _2^2 .
\end{equation}
With the help of $\mathcal{L}_{cyc}^{sp}$ and $\mathcal{L}_{cyc}^{ir}$, we can force the instance-specific encoder and instance-irrelevant encoder to pay more attention to the information they are supposed to encode.

Besides, we also try to explore the temporal information contained in videos to improve the quality of generated samples. Therefore, we follow~\cite{snip} and employ a two-layers bidirectional LSTMs model ($E_t$) to capture the temporal information for model training. The temporal information in face videos is high related to instance-irrelevant features. Thus, the temporal information in the generated pseudo videos is required to be similar to the input one that provides the instance-irrelevant features. We formulate the cycle consistency loss about temporal information as follow:

\begin{eqnarray}
\mathcal{L}_{cyc}^{t} &=& \Vert E_t(K^A) - E_t(T^{B|A}) \Vert _2^2 + \nonumber \\
                      &\;& \Vert E_t(K^B) - E_t(T^{A|B}) \Vert _2^2 \label{eq} ,
\end{eqnarray}
where, the $E_t$ represent the LSTMs model; $K^A$, $K^B$ are the input face video; and the $T^{A|B}$, $T^{B|A}$ denote the generated pseudo videos.

\subsection{Training Objective}
We train our model on the VoxCeleb datasets~\cite{vox1, vox2}, which are large-scale human face videos database. Each video in VoxCeleb only contains one talking person. Then, we optimize our model by the following object:

\begin{eqnarray}
\mathcal{L}_{total} &=& \lambda_{id} \mathcal{L}_{id} + \lambda_{rec} \mathcal{L}_{rec} + \lambda_{orth} \mathcal{L}_{orth} \nonumber \\ &\;
& + \lambda_{cyc} ( \mathcal{L}_{cyc}^{sp} + \mathcal{L}_{cyc}^{ir} + \mathcal{L}_{cyc}^{t}),
\end{eqnarray}

where the $\lambda_{id}$, $\lambda_{rec}$, $\lambda_{orth}$, and $\lambda_{cyc}$ are the hyper-parameters, which is used to control the important of different items.

\begin{table}[t]
\small
\setlength{\tabcolsep}{4.5pt}
\caption{Facial landmark detection performance (NME) on 300W, MAFL, and AFLW datasets. Lower is better.}
\label{tab:fld}

\centering
\begin{tabular}{l c c c}
\toprule
\textbf{Method} & \textbf{300W} & \textbf{MAFL} & \textbf{AFLW} \\
\midrule
\multicolumn{4}{l}{\textbf{Supervised Learning}}\\
\midrule
DCNN & - & 9.73 & 8.97 \\
MTCNN & - & \textbf{5.39} & 6.90 \\
TCDCN & 5.54 & 7.95 & 7.65 \\
Wing & 4.04 & - & \textbf{1.47} \\
SAN & \textbf{3.98} & - & 1.91 \\
\midrule
\multicolumn{4}{l}{\textbf{Unsupervised Learning}}\\
\midrule
\textbf{\textit{Training on CelebA}} & & & \\

UEmbed & 9.30 & 6.67 & 10.53 \\

UDense & 7.97 & 5.83 & 8.80 \\
UDisc & - & 3.16 & \textbf{6.58} \\
Part & - & 3.24 & \\

DVE & 5.75 & 3.42 & 8.60 \\

\midrule
\textbf{\textit{Training on VoxCeleb}} & & & \\
FabNet & 5.71 & 3.44 & 7.52 \\

UCond & - & 3.63 & 6.75 \\
UTrans & 5.37 & - & - \\
\midrule
AIO-Face & \textbf{4.68} & \textbf{2.70} & 6.60 \\
\bottomrule
\end{tabular}

\end{table}

\begin{figure*}[t]
\centering
\includegraphics[width=\textwidth]{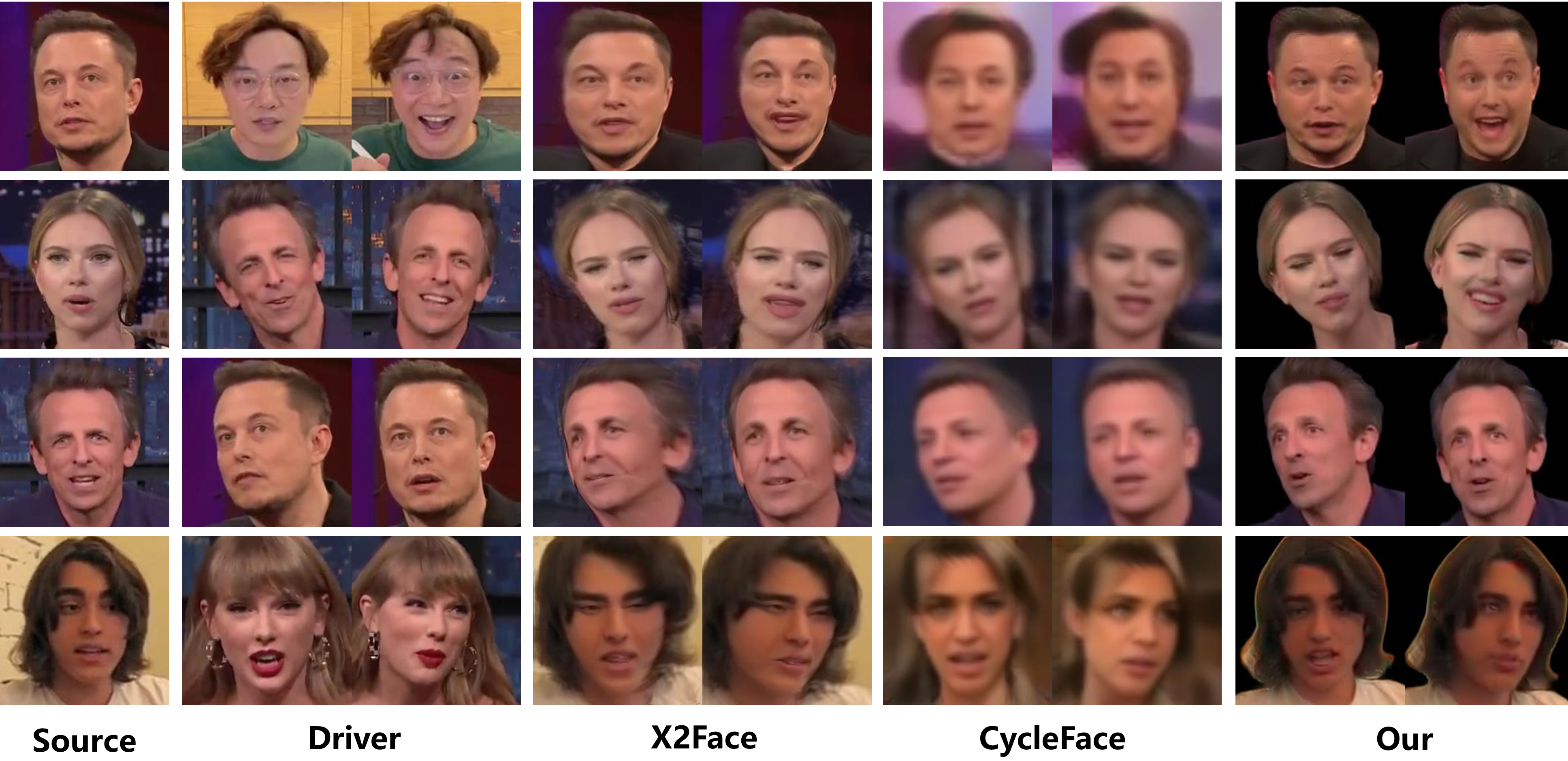}

\caption{Qualitative comparison for cross-person reenactment with state-of-the-art methods including X2Face and CycleFace. The 1-st column is source images, and the 2-nd column is driver images. The 3-rd column demonstrates the results of X2Face, and CycleFace generates the results in the 4-st column. Our method results are illustrated in the 5-st column.}
\label{fig:image_translation}
\end{figure*}

\section{Experiments}
In this section, we firstly introduce the details about the implementation and datasets. Then, we give experimental results on three face analysis tasks: facial landmark detection, face recognition, and facial AU detection. We compare the performance of our method with the state-of-the-art approaches under both the supervised and unsupervised paradigms. Finally, we evaluate the ability of our model in the cross-person reenactment. We also do ablation studies to verify the effectiveness of instance-specific features and instance-irrelevant features on different tasks.

\subsection{Implementation Details}
\textbf{Network architectures.} Our model takes two RGB videos as inputs, and each frame in these videos is resized to $3 \times 256 \times 256$. We employ the ResNet50~\cite{resnet} and MobileNetV2~\cite{mobilenet} as the instance-specific encoder and instance-irrelevant encoder, respectively. The instance-specific encoder encodes the facial representations related to facial recognition with the size of $512 \times 4 \times 4$. And the instance-irrelevant features ($256 \times 4 \times 4$) can be learned by instance-irrelevant encoder.
The target generation module consists of six up-sampling residual blocks. We feed a tensor ($768 \times 4 \times 4$) into the target generate module, then obtain the final results with the size of $3 \times 256 \times 256$.

\textbf{Training data.}
We train our model based on the combination of VoxCeleb1~\cite{vox1} and VoxCeleb2~\cite{vox2} datasets, which are large-scale face videos datasets collected from YouTube. VoxCeleb1 contains 153,516 videos of 1,251 different celebrities, and VoxCeleb2 has 145,569 videos of 5,994 speakers. We select 1 sample every 60 frames from each video and finally get about five million training images.

\textbf{Training strategy.} Training our framework for 600 epochs with a minibatch of 32. The proposed method is implemented with PyTorch. We adopt the AdamW optimizer with $\beta_1=0.5$ and $\beta_2=0.999$. The learning ate is set to 0.0001 and training on four Nvidia A100 GPUs.

\subsection{Evaluations on facial landmark detection}

In the experiments, we introduce a simple regressor including two linear layers for face alignment. We use instance-specific features and instance-irrelevant features extracted by the AIO-Face as the regressor input and train the model on several benchmark datasets. We use the Normalized Mean Error (\textbf{NME}) to evaluate the performance of the facial landmark detector.

We compare the proposed method with the state-of-the-art unsupervised approaches and supervised methods. Table~\ref{tab:fld} reports the results of these methods on 300W~\cite{300w}, MAFL~\cite{mtcnn}, and AFLW~\cite{aflw} datasets, respectively. In particular, some unsupervised learning methods~\cite{un-embedd, un-dense, un-discov, joint_landmark, part_landmark, dve} pre-train on CelebA dataset, and other~\cite{fabnet, un-cond, un-trans} conduct experiments on VoxCeleb dataset.

On the 300W and MAFL datasets, our model achieves 4.68\% and 2.70\% in terms of the NME, which significantly outperforms the evaluated unsupervised baselines by a large margin. Comparing the results on the AFLW dataset, our method is slightly worse than the work of UDisc~\cite{un-discov} because it adopts a more powerful backbone architecture for feature learning. Our proposed method perform better than some supervised approaches, such as MTCNN~\cite{mtcnn}, TCDCN~\cite{tcdnn}, \textit{etc}. But, our method lags behind some state-of-the-art methods~\cite{san, wingloss} in the supervised learning manner.

\begin{table}[t]
\small
\setlength{\tabcolsep}{10pt}
\caption{Face recognition performance on LFW and CPLFW datasets. Higher is better.}
\label{tab:fr}

\centering

\begin{tabular}{l c c}
\toprule
\textbf{Method} & \textbf{LFW} & \textbf{CPLFW} \\
\midrule
\multicolumn{3}{l}{\textbf{Supervised Learning}}\\
\midrule
VGG-Face & 98.95 & 84.00 \\
SphereFace & 99.42 & 81.40 \\
Arc-Face & \textbf{99.53} & \textbf{92.08} \\
\midrule
\multicolumn{3}{l}{\textbf{Unsupervised Learning}}\\
\midrule
LBP & 72.44 & - \\
VGG & 72.20 & - \\
MoCo & 65.88 & 57.82 \\
SimCLR & 75.87 & 62.25 \\
FaceCycle & 73.72 & 58.52 \\
\midrule
AIO-Face & \textbf{76.57} & \textbf{63.05} \\
\bottomrule
\end{tabular}

\end{table}

\begin{table*}[t]
\small
\setlength{\tabcolsep}{4.5pt}
\caption{Facial AU detection performance (F1-score) on DISFA dataset. Higher is better.}
\label{tab:au-disfa}
\centering
\begin{tabular}{l c c c c c c c c c }
\toprule
\textbf{Method} & \textbf{1} & \textbf{2} & \textbf{4} & \textbf{6} & \textbf{9} & \textbf{12} & \textbf{25} & \textbf{26} & \textbf{AVE} \\
\midrule
\multicolumn{10}{l}{\textbf{Supervised Learning}}\\
\midrule

ResNet-Face & 24.9 & 17.9 & 49.5 & 41.2 & 26.2 & 48.6 & 56.4 & 32.8 & 37.2 \\
VGG-Face & 35.5 & 25.5 & 58.1 & 53.8 & 32.4 & 74.4 & 79.0 & 55.7 & 51.8 \\

EAC-Net & 41.5 & 26.4 & \textbf{66.4} & \textbf{50.7} & \textbf{80.5} & \textbf{89.3} & \textbf{88.9} & 15.6 & 48.5 \\
JAANet & \textbf{43.7} & \textbf{46.2} & 56.0 & 41.4 & 44.7 & 69.6 & 88.3 & \textbf{58.4} & \textbf{56.0} \\
\midrule
\multicolumn{10}{l}{\textbf{Unsupervised Learning}}\\
\midrule
SplitBrain & 13.1 & 10.6 & 35.7 & 40.2 & \textbf{30.2} & 57.5 & 77.4 & 40.3 & 38.1 \\
DAE & 17.6 & 12.3 & 46.7 & 43.5 & 26.0 & 62.7 & 64.8 & 47.6 & 40.1 \\
FabNet & 15.5 & 16.2 & 43.2 & \textbf{50.4} & 23.2 & 69.6 & 72.4 & 42.4 & 41.6 \\
TCAE & 15.1 & 15.2 & 50.5 & 48.7 & 23.3 & 72.1 & 82.1 & \textbf{52.9} & \textbf{45.0} \\
Facecycle & 26.4 & 10.2 & 37.3 & 21.5 & 25.0 & 71.8 & \textbf{84.2} & 34.7 & 38.9 \\
\midrule
AIO-Face & \textbf{29.8} & \textbf{19.1} & \textbf{51.2} & 49.0 & 24.0 & \textbf{72.5} & 74.6 & 49.3 & \textbf{46.2} \\
\bottomrule
\end{tabular}

\end{table*}

\subsection{Evaluations on face recognition}
To evaluate the model performance in face recognition, we conduct experiments on LFW~\cite{lfw} and CPLFW~\cite{cplfw} datasets. We employ $E_{id}$ to extract the instance-specific representations from the input samples, then compute the cosine similarity between facial representations and the corresponding ground truth. The results are shown in Table~\ref{tab:fr}. Compared with unsupervised learning methods such as MoCO~\cite{moco}, CycleFace~\cite{facecycle}, and SimCLR~\cite{simclr}, our method shows clear advantages in accuracy. However, it still lags behind supervised methods~\cite{arcface, vggface} significantly in accuracy.

\subsection{Evaluations on facial AU detection}
We investigate the effectiveness of learned instance-irrelevant features by evaluating them on facial AU detection. We introduce a simple classifier with two linear layers follows $E{ir}$ and train them on DISFA~\cite{disfa} datasets. The F1-score is used to measure the performance of the proposed classifier.

Table~\ref{tab:au-disfa} report the performance comparisons DISFA datasets. The results demonstrate that our proposed method outperforms most of unsupervised learning methods~\cite{facecycle, fabnet, tcae} in the average F1 score. Then, we compare the AIO-Face with supervised learning methods, such as VGG-Face~\cite{vggface}, EAC-Net~\cite{eacnet}, and JAA-Net~\cite{jaanet}. However, our method lags behind JAA-Net and EAC-Net, which are both supervised approaches and employ facial landmarks to build patch-specific encoders. These results illustrate that the disentangled instance-irrelevant features contain rich information about facial expressions.

\begin{table}[t]
\small
\setlength{\tabcolsep}{5pt}
\caption{The influence of different features on downstream face analysis tasks. We apply the LFW, DISFA, and 300W for face recognition, facial AU detection, and facial landmark detection, respectively.}
\label{tab:abs_all}

\centering
\begin{tabular}{l c c c}
\toprule
\textbf{Type} & \textbf{$F_{sp}$} & \textbf{$F_{ir}$} & \textbf{$F_{sp}$ + $F_{ir}$}\\
\midrule

Face Rec. & \textbf{76.57} & 35.08 & 73.89 \\
AU Det. & 27.5 & \textbf{46.2} & 41.0 \\
Land. Det. & 6.01 & 10.25 & \textbf{4.68} \\
\bottomrule
\end{tabular}

\end{table}

\section{Ablation Study}
In this section, we first analyze the impact of the instance-specific features and instance-irrelevant features on different downstream tasks in face analysis and conduct extensive experiments to evaluate the contribution of these learned representations. Besides, we also apply our proposed method to cross-person reenactment and compared with state-of-the-art systems.

\subsection{Impact of representations}

In this section, we conduct extensive experiments to evaluate the effectiveness of these learned representations for various downstream tasks in face analysis.

We conduct experiments to evaluate the performance of different features combinations on different downstream tasks including facial landmark detection (Land. Det.), face recognition (Face Rec.) and facial AU detection (AU Det.). The results are shown in Table~\ref{tab:abs_all}. As we can see, head pose estimation and facial AU detection are highly related with face pose and attributes. Therefore, only selecting the instance-irrelevant features achieves the best accuracy. Features required for face recognition are unrelated to expression and pose, so using the instance-specific feature can achieve better performance. Since facial landmark detection is closely related to both instance-specific and instance-irrelevant information such as expression, pose, and identity. Therefore, a more accurate result can be obtained by employing the combined features for facial landmark localization.

\subsection{Reenactment performance}
The proposed AIO-Face can be applied not only to several downstream tasks in face analysis but also for face reenactment. We compare our proposed method with several reenactment systems under the cross-person setting, including X2Face~\cite{x2face} and FaceCycle~\cite{facecycle}. We do not have any ground truth videos for evaluating the model performance for the cross-person reenactment. Therefore, we calculate the cosine similarity (CSIM) between embedding vectors of generated frames and the driver images by pre-trained face recognition model~\cite{arcface}. The CSIM for X2Face and FaceCycle are 0.4089 and 0.3864, respectively. Notable, our AIO-Face achieves 0.5839 in terms of CSIM, significantly outperforming existing methods. Fig.~\ref{fig:image_translation} shows the qualitative comparison results about AIO-Face, X2Face and FaceCycle. Our method can more faithfully reproduce the head pose and expression from the driver frames without noticeable artifacts.

\section{Conclusion}
This paper presents AIO-Face, a unified model based on the unsupervised learning framework, which effectively addresses multiple face analysis tasks. Our method learns to disentangle the instance-specific and instance-irrelevant features from a large collection of unlabeled face videos. These disentangled features are then utilized for various facial analysis tasks, including facial landmark detection, face recognition, and facial AU detection. To achieve effective feature disentanglement, our approach incorporates a feature disentanglement module, a target generation module, and a cycle-consistency regularization module. The feature disentanglement module consists of two encoders, one for extracting the instance-specific features and the other for instance-irrelevant features. The target generation module aims to reconstruct the videos based on the disentangled features. The cycle-consistency regularization employs a cross-verified disentangling strategy to ensure that the latent features remain consistent with the original one, providing additional supervision during model training. Extensive experiments demonstrate the superior performance of our method compared to the evaluated unsupervised learning baselines and even comparable to some supervised methods.

\clearpage
{
  \small
  \bibliographystyle{ieeenat_fullname_citation}
  \bibliography{references}
}
\clearpage
\appendix
\setcounter{section}{0}
\setcounter{figure}{0}
\setcounter{table}{0}
\setcounter{equation}{0}
\renewcommand{\thefigure}{S\arabic{figure}}
\renewcommand{\thetable}{S\arabic{table}}
\renewcommand{\theequation}{S\arabic{equation}}
\renewcommand{\theHfigure}{S\arabic{figure}}
\renewcommand{\theHtable}{S\arabic{table}}
\renewcommand{\theHequation}{S\arabic{equation}}
\maketitlesupplementary
\section{Experiments}
In this section, we report some additional results of AIO-Face on downstream tasks, including head pose estimation, facial expression recognition, facial AU detection, and facial landmark detection.

\subsection{Evaluations on head pose estimation}
As our assumption, the information about head pose estimation should be encoded in the instance-irrelevant features. Therefore, we investigate the effectiveness of instance-irrelevant features by evaluating their performance on head pose regression. Similar to the facial landmark detection, we only train a simple linear regressor for head pose estimation, which tries to predict three head pose angles, including pitch, yaw, and roll. Specifically, the linear regressor only takes the instance-irrelevant features as input and maps them to three head pose angles.

\begin{table}[t]
\small
\setlength{\tabcolsep}{3pt}
\caption{Head pose estimation performance on AFLW2000 datasets. Lower is better.}
\label{tab:hpe}

\centering
\begin{tabular}{l c c c c}
\toprule
\textbf{Method} & \textbf{Roll} & \textbf{Pitch} & \textbf{Yaw} & \textbf{MAE} \\
\midrule
\multicolumn{5}{l}{\textbf{Supervised Learning}}\\
\midrule

Dlib~\cite{dlib} & 10.50 & 13.60 & 23.10 & 15.80 \\
FAN~\cite{fan} & 8.71 & 12.3 & 6.36 & 9.12 \\
FSA~\cite{fsa-net} & 5.28 & 6.71 & 5.27 & 5.75 \\
\midrule
\multicolumn{5}{l}{\textbf{Unsupervised Learning}}\\
\midrule
TCAE~\cite{tcae} & 14.83 & 14.57 & 21.75 & 17.39 \\
MoCo~\cite{moco} & 15.55 & 16.29 & 28.49 & 20.11 \\
SimCLR~\cite{simclr} & 12.08 & 19.86 & 11.20 & 14.38 \\
FabNet~\cite{fabnet} & 14.51 & 13.25 & 13.92 & 13.89 \\
FaceCycle~\cite{facecycle} &  12.94 & \textbf{12.76} & 11.70 & 12.47 \\
PCL~\cite{pcl} & 10.62 & 16.59 & \textbf{9.86} & 12.36 \\
\midrule
AIO-Face & \textbf{10.57} & 13.05 & 10.24 & \textbf{11.28} \\
\bottomrule
\end{tabular}

\end{table}

\begin{table}[t]
\small
\setlength{\tabcolsep}{4.5pt}
\caption{Expression recognition performance on FER-2013 and RAF-BD datasets in terms of accuracy (\%). Higher is better.}
\label{tab:fer}

\centering
\begin{tabular}{l c c}
\toprule
\textbf{Method} & \textbf{FER-2013} & \textbf{RAF-BD}  \\
\midrule
\multicolumn{3}{l}{\textbf{Supervised Learning}}\\
\midrule
FSN~\cite{fsn} & 67.60 & 81.10 \\
ALT~\cite{alt} & 69.85 & 84.50 \\
\midrule
\multicolumn{3}{l}{\textbf{Unsupervised Learning}}\\
\midrule
FabNet~\cite{fabnet} & 46.98 & 66.72 \\
TCAE~\cite{tcae} & 45.05 & 65.32 \\
RankAU~\cite{rankau} & 47.61 & 68.32 \\
SimCLR~\cite{simclr} & 49.51 & 71.06 \\
MoCo~\cite{moco} & 47.24 & 68.32 \\
FaceCycle~\cite{facecycle} & 48.76 & 71.01 \\
\midrule
AIO-Face & \textbf{50.15} & \textbf{72.29} \\
\bottomrule
\end{tabular}

\end{table}

\begin{table*}[t]
\small
\setlength{\tabcolsep}{4pt}
\caption{Facial AU detection performance (F1-score) on BP4D dataset. Higher is better.}
\label{tab:au-bp4d}
\centering
\begin{tabular}{l c c c c c c c c c c c c c}
\toprule
\textbf{Method} & \textbf{1} & \textbf{2} & \textbf{4} & \textbf{6} & \textbf{7} & \textbf{10} & \textbf{12} & \textbf{14} & \textbf{15} & \textbf{17} & \textbf{23} & \textbf{24} & \textbf{AVE} \\
\midrule
\multicolumn{14}{l}{\textbf{Supervised Learning}}\\
\midrule

ResNet-Face~\cite{resnet} & 39.3 & 40.6 & 38.5 & 64.2 & 67.5 & 71.0 & 65.3 & 57.2 & 37.8 & 51.3 & 35.1 & 32.6 & 49.9 \\
VGG-Face~\cite{vggface} & 46.4 & 36.3 & 49.6 & 76.0 & 77.6 & 80.2 & 87.8 & 60.8 & 40.4 & 59.1 & 43.7 & 48.2 & 58.8 \\
AlexNet~\cite{alexnet} & 40.3 & 39.0 & 41.7 & 62.8 & 54.2 & 75.1 & 78.1 & 44.7 & 32.9 & 47.3 & 27.3 & 40.1 & 48.6 \\
DRML~\cite{drml} & 36.4 & 41.8 & 43.0 & 55.0 & 67.0 & 66.3 & 65.8 & 54.1 & 33.2 & 48.0 & 31.7 & 30.0 & 48.3 \\
EAC-Net~\cite{eacnet} & 39.0 & 35.2 & 48.6 & 76.1 & 72.9 & 81.9 & 86.2 & 58.8 & 37.5 & 59.1 & 35.9 & 35.8 & 55.9  \\
JAANet~\cite{jaanet} & 47.2 & 44.0 & 54.9 & 77.5 & 74.6 & 84.0 & 86.9 & 61.9 & 43.6 & 60.3 & 42.7 & 41.9 & 60.0 \\
FAT~\cite{fat} & 51.7 & \textbf{49.3} & \textbf{61.0} & 77.8 & \textbf{79.5} & 82.9 & 86.3 & \textbf{67.6} & \textbf{51.9} & 63.0 & 43.7 & \textbf{56.3} & \textbf{64.2} \\
PIAP~\cite{piap} & \textbf{54.2} & 47.1 & 54.0 & \textbf{79.0} & 78.2 & \textbf{86.3} & \textbf{89.5} & 66.1 & 49.7 & \textbf{63.2} & \textbf{49.9} & 52.0 & 64.1 \\
\midrule
\multicolumn{14}{l}{\textbf{Unsupervised Learning}}\\
\midrule

DAE~\cite{dae} & 39.5 & 34.5 & 40.8 & 70.5 & 68.4 & 76.3 & 82.9 & 60.7 & 23.1 & 54.1 & 34.3 & 43.1 & 52.3 \\
FabNet~\cite{fabnet} & 43.3 & 35.7 & 41.6 & 72.9 & 63.0 & 75.9 & 83.5 & 57.7 & 26.5 & 48.2 & 33.6 & 42.4 & 52.0 \\
TCAE~\cite{tcae} & 43.1 & 32.2 & 44.4 & 75.1 & 70.5 & 80.8 & 85.5 & 61.8 & \textbf{34.7} & \textbf{58.5} & \textbf{37.2} & \textbf{48.7} & \textbf{56.1} \\
RankAU~\cite{rankau} & 42.3 & 24.3 & 44.1 & 71.8 & 67.8 & 77.6 & 83.3 & 61.2 & 31.6 & 51.6 & 29.8 & 38.6 & 52.0 \\
SimCLR~\cite{simclr} & 11.3 & 6.0 & 19.7 & 67.5 & 72.2 & 81.1 & 75.8 & 52.6 & 16.0 & 22.3 & 4.5 & 9.4 & 36.5 \\
MoCo~\cite{moco} & 7.6 & 2.9 & 13.7 & \textbf{79.2} & \textbf{79.8} & \textbf{85.0} & 87.7 & 61.6 & 33.7 & 56.8 & 16.3 & 28.8 & 46.1 \\
\midrule
AIO-Face & \textbf{44.6} & \textbf{35.9} & \textbf{45.0} & 74.1 & 78.1 & 81.6 & \textbf{88.0} & \textbf{61.9} & 33.5 & 56.4 & 32.5 & 46.9 & \textbf{56.5} \\
\bottomrule
\end{tabular}

\end{table*}

\begin{figure}[t]
\centering
\includegraphics[width=\columnwidth]{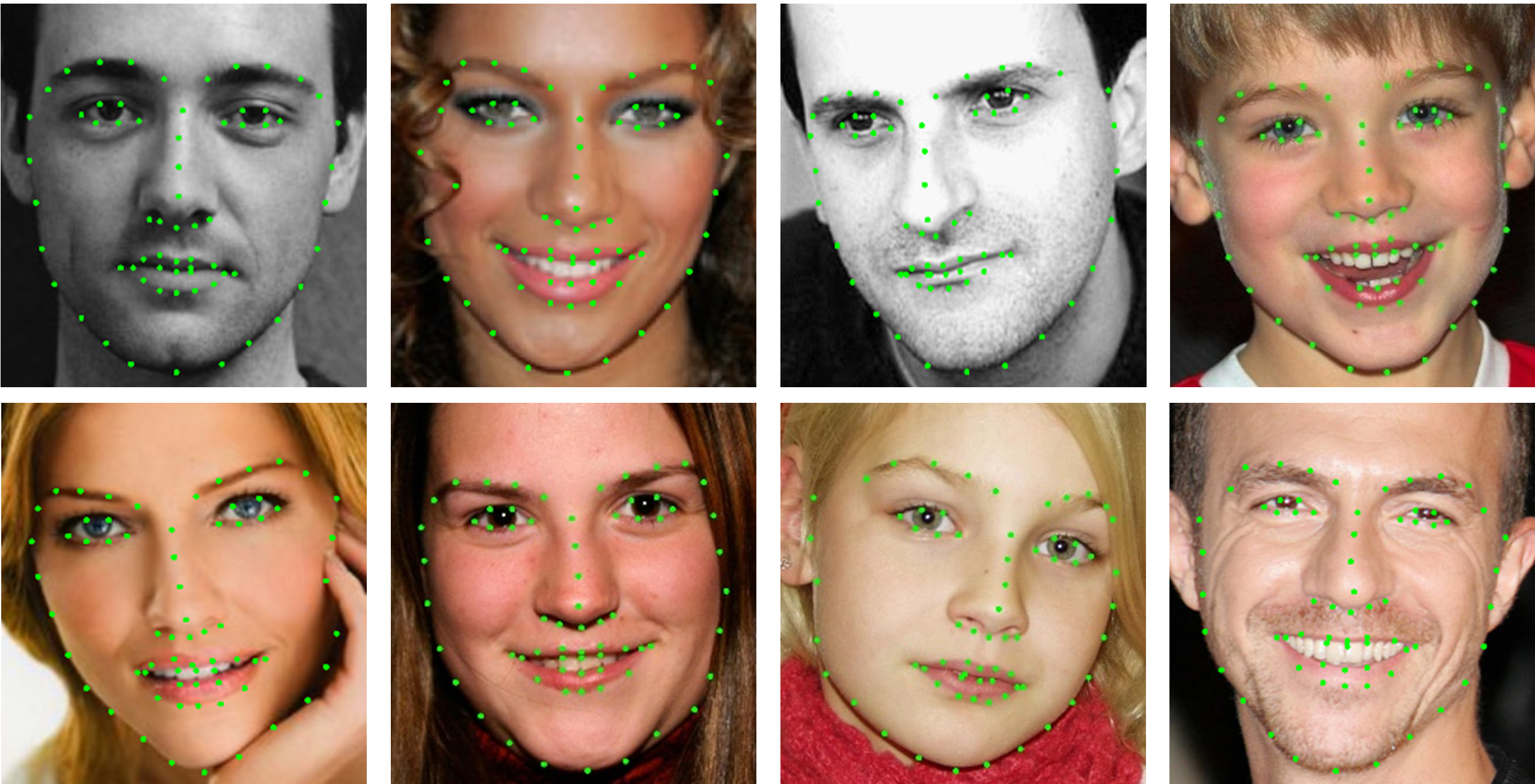}
\caption{Visualization results of the AIO-Face in facial landmark detection. The test images are selected from 300W dataset.}
\label{fig:vis-300w}
\end{figure}

\begin{figure*}[t]
\centering
\includegraphics[width=\textwidth]{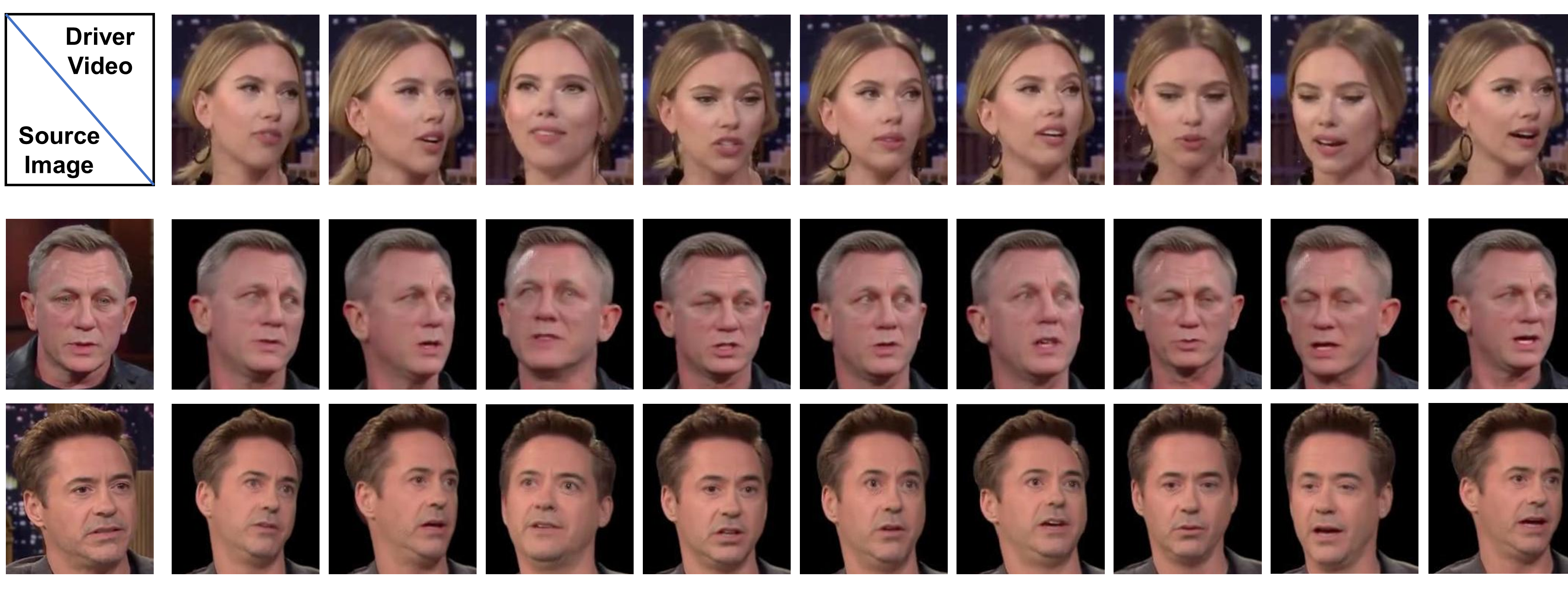}

\caption{Generated images by AIO-Face. The results demonstrate that our method is able to disentangle the instance-specific features and instance-irrelevant features.}
\label{fig:image_translation_video}
\end{figure*}

We evaluate our AIO-Face and the existing methods under the AFLW2000 dataset~\cite{3ddfa}. It contains 2000 face images, which are divided into the training set and validation set. All the face images are cropped around the face then fed to our model. The Mean Absolute Error (\textbf{MAE}) is used as the metric to evaluate the accuracy of head pose estimation, which is defined as follow:
\begin{equation}
MAE(P, P^*) = \frac{1}{N}\sum_{i=1}^{N}(\Vert p_i - p_i^* \Vert),
\end{equation}
where $N$ is the number of face images.
The $p_i$ and $p_i^*$ denote the ground truth and predicted head posture, respectively.

We compare our head pose estimation regressor with the state-of-the-art unsupervised and supervised methods. Table~\ref{tab:hpe} shows the results on AFLW2000 dataset. We can find that the simple regressor based on instance-irrelevant features achieves 11.28\% in MAE. Compared with unsupervised learning approaches (MoCo~\cite{moco}, TCAE~\cite{tcae}, SimCLR~\cite{simclr}, FabNet~\cite{fabnet}, FaceCycle~\cite{facecycle}, PCL~\cite{pcl}), our proposed method has significant advantages in prediction accuracy. Results demonstrate that the instance-irrelevant features learned by the AIO-Face contain enough information used to describe the head posture. However, our method lags behind state-of-the-art methods for head pose estimation such as FAN~\cite{fan} and FSA~\cite{fsa-net}.

\subsection{Evaluations on expression recognition}
We evaluate the performance of AIO-Face in the task of expression recognition on two benchmark datasets, including FER-2013~\cite{fer2013} and RAF-DB~\cite{raf}. The FER-2013 is a public dataset annotated with seven emotions. It comprises over 30,000 gray-scale face images, with 28,709 samples in the training set and 3,589 in the test set. RAF-DB is a large-scale dataset for facial expression recognition that contains around 30,000 different face images. Each sample is annotated with seven basic emotions. Following the setup of~\cite{facecycle}, we select 12,271 face images as the training set and 3,076 as the test set. Similar to other downstream tasks, we train a simple classifier based on instance-irrelevant features for facial expression recognition. The Accuracy is employed to validate the model performance.
The results of expression recognition can be found in Table~\ref{tab:fer}. It is worth noting that our AIO-Face performs better than other unsupervised learning-based methods.

\subsection{Evaluations on facial AU detection}
Facial AU detection is a more challenging task in face analysis. Therefore, we can investigate the effectiveness of learned instance-irrelevant features by evaluating them on facial AU detection. Similar to the task of head pose estimation, the proposed classifier only takes the instance-irrelevant features as input. We report the additional results of our method on BP4D~\cite{bp4d}. It contains 328 videos, including about 140,000 frames with AU labels of occurrence or non-occurrence. There are 8 sessions in BP4D with 23 females and 18 males. Follow the setting of~\cite{tcae}, 12 AUs (\textit{i.e.}, 1, 2, 4, 6, 7, 10, 12, 14, 15, 17, 23, and 24) are used for evaluation.

We use the frame-based F1-score to measure the performance of the proposed classifier. For each facial AU, we get the F1 as follow:
\begin{equation}
F_1 = \frac{2PR}{P + R},
\end{equation}
where the \textit{R} and \textit{P} represent the recall and precision, respectively. For each method that needs to be compared, the average F1 is calculated in terms of all the investigated AUs.

We compare our proposed AIO-Face with the state-of-the-art facial AU detection methods. Table~\ref{tab:au-bp4d} report the performance comparisons on BP4D dataset. Compared with general face analysis methods (\textit{e.g.}, FabNet and DAE), our approach has more advantages in AU detection.

\subsection{Evaluations on facial landmark detection}
We evaluate the proposed facial landmark detector on 300W, and some visualization results are shown in Fig.~\ref{fig:vis-300w}. The results demonstrate that our proposed method can perform well in the task of facial landmark detection.

\subsection{Evaluations on face reenactment}
With the help of the LSTMs model, our method can generate realistic results with temporal continuity. Fig.~\ref{fig:image_translation_video} illustrates that our method makes the pose change smoothly. In addition, it also shows that our method could transfer the instance-irrelevant information from the driver to the source while preserving the instance-specific representations about the source image.

\end{document}